\documentclass[a4paper,fleqn]{cas-sc}

\usepackage[numbers]{natbib}

\usepackage{graphicx}
\usepackage{textcomp}
\usepackage{xcolor}
\usepackage{tikz}
\usepackage{amsthm}
\usepackage{comment}
\usetikzlibrary{shapes.geometric, arrows}
\usepackage[ruled,vlined]{algorithm2e}
\usepackage{parnotes}
\usepackage{url}

\newtheorem{theorem}{Theorem}
\newtheorem*{theorem*}{Theorem}
\newtheorem*{remark}{Remark}

\def\tsc#1{\csdef{#1}{\textsc{\lowercase{#1}}\xspace}}
\tsc{WGM}
\tsc{QE}
\tsc{EP}
\tsc{PMS}
\tsc{BEC}
\tsc{DE}

\begin{document}
\let\WriteBookmarks\relax
\def\floatpagepagefraction{1} 
\def\textpagefraction{.001}
\shorttitle{CDiNN}
\shortauthors{Parameswaran S et~al.}

\title [mode = title]{CDiNN -Convex Difference Neural Networks}                      

%

%


\author[1]{Parameswaran Sankaranarayanan}
\ead{paramesh282@gmail.com}

\address[1]{Robert Bosch Centre for Data Science and Artificial Intelligence, Indian Institute of Technology Madras, Chennai, India}

\author[1]{Raghunathan Rengaswamy}
\cormark[1]
\ead{raghur@mail.iitm.com}

\cortext[cor1]{Corresponding author}


\begin{abstract}
Neural networks with ReLU activation function have been shown to be universal function approximators and learn function mapping as non-smooth functions.  Recently, there is considerable interest in the use of neural networks in applications such as optimal control. It is well-known that optimization involving non-convex, non-smooth functions are computationally intensive and have limited convergence guarantees. Moreover, the choice of optimization hyper-parameters used in gradient descent/ascent significantly affect the quality of the obtained solutions. A new neural network architecture called the Input Convex Neural Networks (ICNNs) learn the output as a convex function of inputs thereby allowing the use of efficient convex optimization methods. Use of ICNNs for determining the input for minimizing output has two major problems: learning of a non-convex function as a convex mapping could result in significant function approximation error, and we also note that the existing representations cannot capture simple dynamic structures like linear time delay systems. We attempt to address the above problems by introduction of a new neural network architecture, which we call the CDiNN, which learns the function as a difference of polyhedral convex functions from data. We also discuss that, in some cases, the optimal input can be obtained from CDiNN through difference of convex optimization with convergence guarantees and that at each iteration, the problem is reduced to a linear programming problem.

\end{abstract}


\begin{highlights}
\item Introduce new Neural network architecture for efficient decision making without significant loss of representational capability
\item Use of Difference of Convex (DC) programming in decision making involving neural networks
\item DC optimization produces better result at each iteration and guarantees convergence with the proposed neural network architecture
\item Optimization problem at each step reduces to Linear Programming problem with the proposed network architecture
\item Illustration of advantages of this neural network structure through several case studies
\end{highlights}


\maketitle

\section{Introduction}
Neural networks have been used for function approximation and classification tasks in several applications. The fact that these networks are shown to be universal function approximators is a nice theoretical guarantee for possible representational capability. Specifically, in \cite{NNUniversalApprox2004}, the authors have shown the universal approximation capability of neural networks with unbounded activation functions like ReLU.  Further, in recent years, complex problems have been solved using deep architectures (multiple layers) with Rectified Linear Unit (ReLU) activation functions. This has spawned a tremendous interest in the study of these architectures across all engineering domains. However, while these properties are theoretically nice, the actual performance depends on solving  optimization problems in multiple contexts while neural networks are used. Optimization is important in neural networks in two contexts: 
\begin{enumerate}
\item The first context is the training of these networks, where the network parameters are identified by minimizing a loss function. The most popular approach is the back-propagation approach, which is a gradient descent algorithm. Several variants of this algorithm have been evaluated for neural network training. Since the loss function is usually non-convex, issues related to convergence and local minima problems need to be addressed. In spite of all these, there are several applications where neural networks have found great success.
\item The second context is where the trained network is used in model-based control or model-based optimization studies. These are problems where optimal decisions have to be made such as understanding the effect of inputs on certain neurons, inputs for optimal control and so on. Since the identified neural network model is non-convex, the use of such models in any control and/or optimization application is beset with local minima problems. Global optimization solutions are computationally prohibitive for use in real-time or near real-time applications. Further, there is very little theoretical convergence guarantees for many of these techniques. An additional difficulty is the trial-and-error process for choosing parameters in the optimization procedure (referred to as hyper-parameter tuning), which leads to considerable fluctuations in the quality of results. This could possibly make the whole process non-standard.
\end{enumerate}

All nonlinear problems are not complex from an optimization perspective, it is the non-convexities that lead to difficulties. On the other hand, algorithms dealing with convex functions have guarantees regarding convergence and identification of global optimum. As a result, use of convex optimization ideas in neural networks is gaining prominence. In \cite{ICNN2017} and \cite{optimalnn}, the authors propose an architecture called the Input Convex Neural Network (ICNN) in which the output is learnt as a convex function of the inputs. When a network like that is used in optimization studies, particularly when the output function is directly optimized, then the optimization problem becomes a  convex optimization problem. As a result, efficient convex optimization algorithms can be applied to solve these types of problems. Further, an extension of the ICNN as a recurrent neural network has also been proposed; this structure can be used to model time series data.

While there are several advantages to ICNN, one of the main drawbacks  of the architecture is the representational capability. There can be considerable loss in accuracy when approximating a non-convex function using a convex function. Additionally, we show that the ICNN architecture, due to the manner in which it is constructed, cannot be used to model time delay systems. To address these difficulties, while retaining some of the advantages in working with convex functions in optimization, we propose a new neural network architecture called the CDiNN architecture. In this architecture, any given function that needs to be modeled is represented as a difference of convex functions. This enhances the representation capability tremendously. Further, as in ICNN, when the outputs are directly optimized, one could use difference of convex algorithms (DCA), which provide much better theoretical guarantees than the standard gradient descent algorithms. In this paper, we describe the CDiNN architecture, analyze its properties and show application examples to demonstrate the versatility of this architecture.  

The rest of the paper is organized as follows. ICNN and their properties, advantages and disadvantages are discussed in the next section. After that we describe the proposed CDiNN architecture, explore its properties and  compare the proposed architecture with ICNN. Application case studies that corroborate the theoretical claims follows this discussion. We conclude with thoughts on further development of the CDiNN architecture.

\section{Input Convex Neural Network - ICNN}

 Input Convex Neural Network (ICNN) \cite{ICNN2017} is designed to learn function mapping from a class of convex functions generated using a neural network architecture. In this neural network structure, outputs are convex functions of inputs. This is achieved through appropriate choices for activation functions and constraining the weights. In this section, we start with a basic description of the ICNN architecture. Properties of ICNN and some of the representational issues in the use of the ICNN architecture are then described. Training strategies for identifying the weights in the ICNN structure are outlined. Finally, applications of ICNN studied in the literature and the disadvantages of the ICNN structure are explored. 
 
 \subsection{ICNN Structure}
\begin{figure}
	\centering
		\includegraphics[scale=.6]{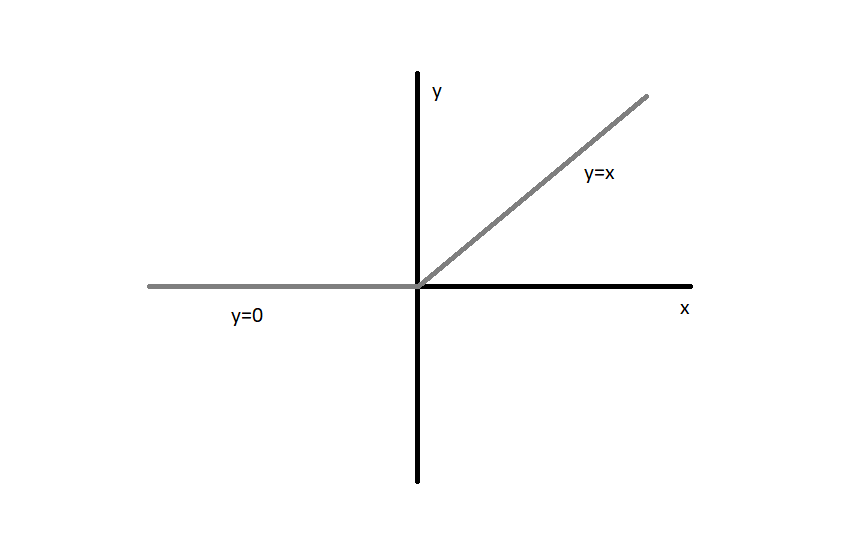}
	\caption{ReLU activation function output $=$ max(0, input)}
	\label{FIG:ReL}
\end{figure}

\begin{figure}
	\centering
		\includegraphics[scale=.75]{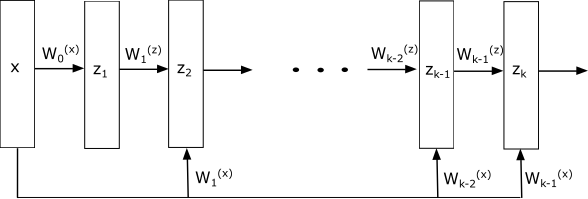}
	\caption{Input Convex Neural Network proposed in \cite{ICNN2017}}
	\label{FIG:1}
\end{figure}

 Input Convex Neural Network is a modification of feed-forward neural networks with constraints on the weights and the choice of activation functions \cite{ICNN2017}. The architecture of the neural network is shown in Figure \ref{FIG:1}. The input to the network ($x$) is passed through hidden layers (possibly deep) and $z_k$ is the output of the $k^{th}$ layer. The activation function used in the network is a convex non-decreasing function. Rectified Linear Unit (ReLU), shown in Figure \ref{FIG:ReL} is an example of a possible activation function that can be used. The input $x$ is connected to hidden layers through weights $W_i^{(x)}$ and output of hidden layer is connected to the next layer through weights $W_i^{(z)}$. The network structure depicted in Figure \ref{FIG:1} reduces to the standard feed-forward structure, except for the direct connections from the inputs to every layer of the network.   It can be observed that when the weights connecting the output of one hidden layer of neural network to the next layer ($W_i^{(z)}$) is constrained to be positive and the activation function is convex non-decreasing function, every element in the output layer ($z_K$) is a convex function of the input $x$. This is because a non-negative sum of convex functions is a convex function and a composition of non-decreasing convex function $g_i$ on a convex function results in a convex function. Given that the weights are constrained to be positive, the pass-through layers (connections from the inputs directly to every layer) provided by $W_i^{(x)}$ are required to permit identity mapping.

\begin{equation}
\label{eq1}
z_{i+1} = g_i(W_i^{(z)} z_i + W_i^{(x)} x + b_i);
\end{equation}

\begin{figure}
	\centering
		\includegraphics[scale=.75]{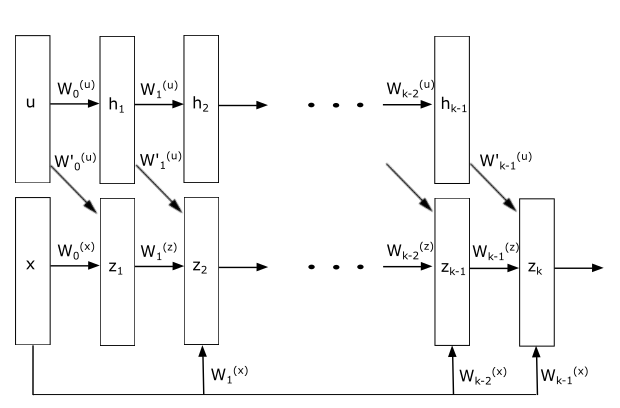}
	\caption{Partial Input Convex Neural Network}
	\label{FIG:1p} 
\end{figure}

It is also possible easily extend this framework to construct a Partial Input Convex Neural Network (PICNN) where the output is a convex function of only a subset of the input terms. In the Figure \ref{FIG:1p}, each element of $z_k$ can be learnt as a convex function of x and a non-convex function of u. Only the weights $W_i^{(z)}$ are constrained to be positive.

\subsection{Recurrent ICNN}
\begin{figure}
	\centering
		\includegraphics[scale=.75]{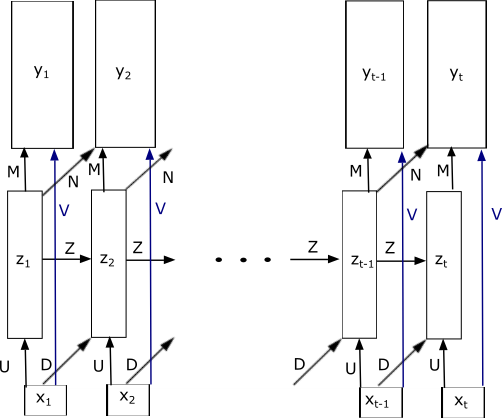}
	\caption{Recurrent Input Convex Neural Network expanded in time (time=1 to time=t)}
	\label{FIG:ricnn} 
\end{figure}

We discuss the Recurrent ICNN architecture proposed in \cite{optimalnn}. The architecture of Recurrent ICNN is shown in Figure \ref{FIG:ricnn} and described by equation \ref{eq_icr} 
\begin{eqnarray}
	 z_t & = & ReLU( U x_t +  Z  z_{t-1} +  D x_{t-1}) \nonumber \\
	y_t & = & f_a( M  z_t +  N  z_{t-1} +  V x_t) \label{eq_icr}
\end{eqnarray}

where $ x_t$ is the input vector at time t in the format $x = \begin{bmatrix}  u \\- u\end{bmatrix}$, u is the input, $y_t$ is the output of the neural network at time t, $f_a$ is a non-decreasing convex activation function (Linear, ReLU etc.) and $D, N, V$ are the weights of pass-through layers. It can be observed that the output $y_t =f(x_1, x_2, ..., x_t; \theta)$ is a convex function of the inputs. $\theta = {[U, V, M, N, D, Z]}$ are network parameters and are constrained to be non-negative in this formulation.

\subsection{Properties}

The authors of \cite{optimalnn} compare the efficiency of representation for a convex function $f$ as ICNN with piece-wise affine representation. It is argued that input convex neural networks provide a compact approach to learn a function as a maximum of affine functions.

\begin{theorem}
\label{th2}
For any Lipschitz convex function over a compact domain, there exists an input convex neural network with non-negative weights (of $W_i^{(z)}$) and ReLU activation functions that approximates it within some $\epsilon$. \cite{optimalnn}
\end{theorem}

\begin{remark}
Any Lipschitz convex function can be approximated by a maximum of finite number of affine functions.The proof of the above theorem involves showing that a convex function in the form of maximum of $k$ affine functions can be represented by ICNN with $k$ hidden layers, each with one neuron and ReLU activation function \cite{optimalnn} . 
\label{themax2}
\end{remark} 

\subsection{Learning of the network parameters}
Two methods for learning network parameters are described in \cite{ICNN2017}. 
\begin{enumerate}
    \item \textbf{Direct functional fitting - Gradient descent}:  Gradient descent (or its variants) are used to minimize a loss function - for example, mean square error. At each iteration, the weights $W_i^{(z)}$ are forced to be positive by replacing the negative weights by $0$ or multiplication with $-0.5$.
    \item \textbf{Argmin differentiation}: In this method, the authors propose to learn the network parameters by minimizing the difference between true optimal input and the estimated input that minimizes the output of the network. This is suitable for cases where the true optimal input is known at the time of training. 
    \end{enumerate}
    
\subsection{Application of ICNN to optimal control}
In  \cite{optimalnn}, the authors discuss two configurations to use ICNN in optimal control. In the first configuration, the ICNN is used to learn the state transition as a function of actions and states in reinforcement learning in feedforward configuration. In the second configuration, the authors used Input Convex Recurrent Neural Network (ICRNN) for modelling the cost as a dynamic function of inputs and states. The authors then show results for a model predictive controller that minimizes the positively weighted sum of cost functions (ICRNNs) with respect to the inputs $x$.

In \cite{YANG2021107143}, the authors used ICNN for model predictive control of Van de Vusse reaction. The authors observe that ICNN does not perform well in modelling the reaction (with a large MSE); this reiterates the fact that the choice of ICNN for modeling nonconvex functions can be problematic. However, in this study it is shown that the control performance with ICNN is reasonably good for this problem. Nonetheless,  it is not possible to guarantee that this can be generalized for other problems. The authors argue that given the predictability of results and no local optima points, ICNN could be a good choice for modelling provided the model-mismatch is not significant enough to generate infeasible controllers. A few other examples where ICNN is used for optimal control are  \cite{bunning2020input} and \cite{chen2020input}, where ICNN is applied for room temperature control and optimal voltage regulation respectively.

\subsection{Discussions}
In this section, we discuss the disadvantages of using ICNN in decision making process by the use of simple functions. We also show that Recurrent ICNN cannot represent time delay systems.
\subsubsection{1D non-convex functions}
\begin{figure}
	\centering
		\includegraphics[scale=.4]{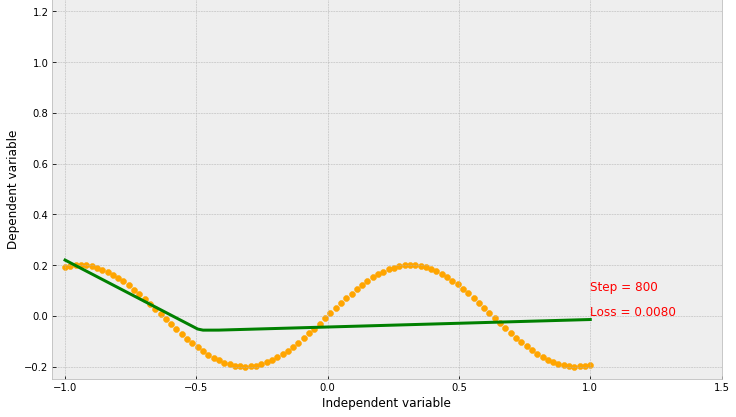}
	\caption{Learning of sine function by ICNN of one hidden layer and 10 hidden neurons. The green curve is the learnt function, the yellow points represent the training data. Mean square Loss of training data is 0.008 and number of training epochs  is 800.}
	\label{FIG:a}
\end{figure}

While ICNNs provide an interesting framework for using neural networks in optimization, we note that many interesting optimization problems involve non-convex functions and in such scenarios, convex function approximations could lead to significant errors. Consider the use of ICNN for approximating a non-convex function $y = sin(5x)/5$ where, $x \in [-1, 1]$. The approximation results are shown in Figure \ref{FIG:a} and two important observations can be made.
\begin{enumerate}
    \item Consider optimization involving ICNN to find minima with initial point as x=0.5. From the Figure \ref{FIG:a}, we see that optimizing using ICNN would result in an optimal value for $x=-0.5$, while $x=1.0$ is the closest best optimum. Hence it could be observed that if ICNN is used for control, it might result in optimization solutions that could be very far away from the present operating region (while skipping an equivalent solution that might be nearby).
    \item The global optimum value pointed by the ICNN in Figure \ref{FIG:a} is $x = -0.5$ whereas the true global optima occurs at $x=-0.314$. Hence we see that global optimum found by ICNN may not be the true optimum value.
\end{enumerate}

\subsubsection{Modelling time delay systems}
\label{tds1}
The recurrent ICNN architecture is empirically shown to model dynamical systems well in \cite{optimalnn}. We now study this characteristic of ICNN through an example.  Let us consider modelling of $y_t = u_{t-5}$ for illustration. We focus on the ability of the Recurrent ICNN shown in Figure \ref{FIG:ricnn} to represent the negative values of $u_{t-5}$. Consider $t=5$ for the analysis. Only at time steps $t=0$ and $t=1$, the hidden layers receive the input $u_{0}$ directly. Further, we note that the output of the hidden layers are non-negative because ReLU clips negative values. Hence it would be difficult to model the negative values of $u_{0}$ since $y_5$ is connected to $u_{0}$ only via hidden layers. Interestingly, we observe that, in the presence of bias and bounded inputs $u$, this function can be learnt by the network due to the following. With enough hidden nodes to allow for dynamical mapping, the network could map $u_{0}$ to $y_5$ by shifting all the inputs in the input layer to positive (by a bias of $b$) and shifting back the values at output layer to input values by a bias of $-b$, thereby preserving the information.

\begin{figure}
	\centering
		\includegraphics[scale=.3]{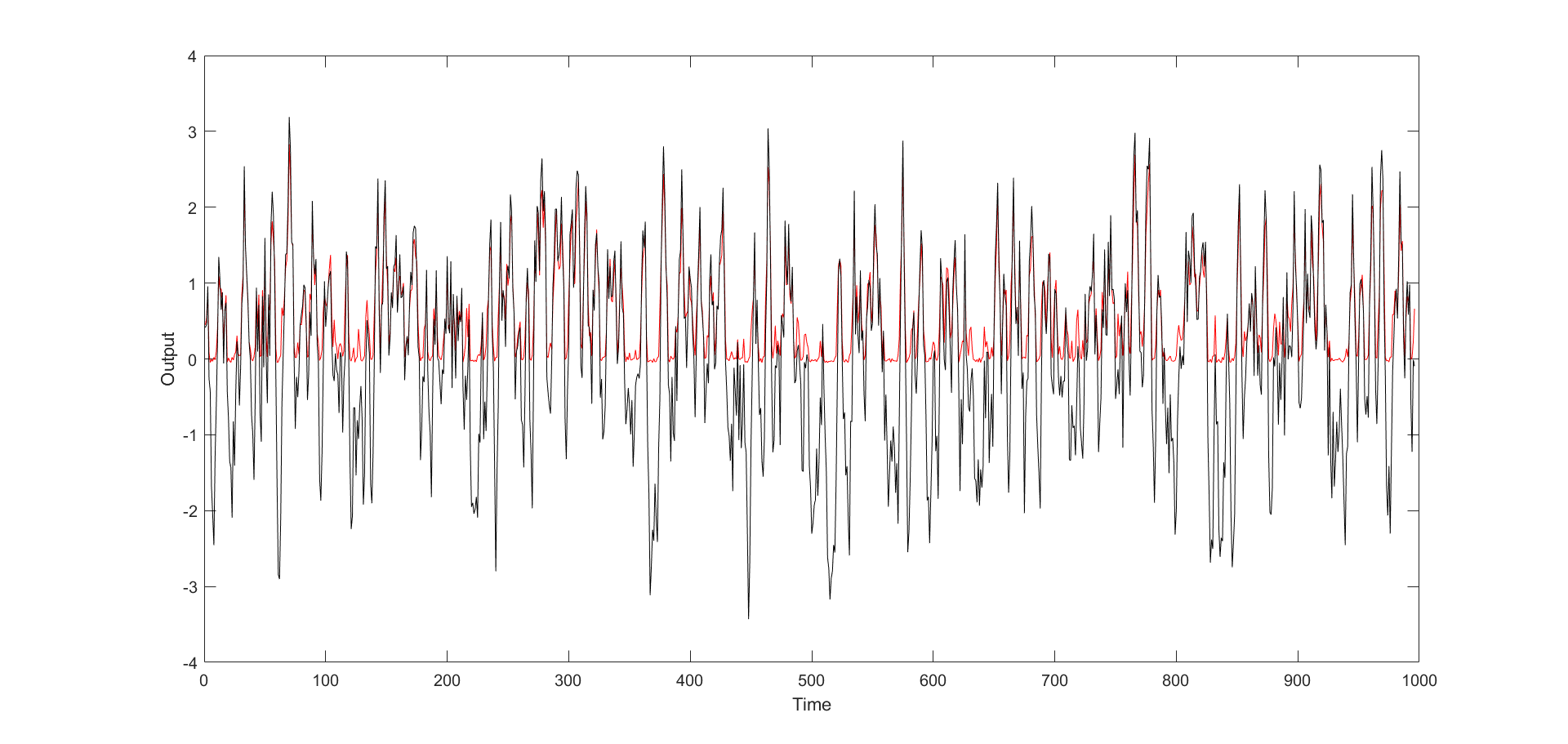}
	\caption{Output predicted (red) vs truth (black) by Recurrent ICNN (without bias) as proposed in \cite{optimalnn}}
	\label{FIG:b}
\end{figure}

Thus, we investigate if this network could still learn the time-delay functions in absence of bias. We show the results empirically with Recurrent ICNN with 10 recurrent neurons and disabled biases. The input that is used to excite the system is a random uniform signal uniformly sampled from [-1,1]. The output is $y_{t} = u_{t-4} + u_{t-3} + u_{t-2} + u_{t-1}$. There is no noise in the output signal. The input to the RNN is time delayed input signal with delay $\in [0,4]$. It is seen from the Figure \ref{FIG:b} that ICNN does not capture the dynamic behaviour of this system. Hence, we attribute the performance of Recurrent ICNN in time-delay systems to shifting of inputs by biases. While this is acceptable in training data, this could result in poor generalization when the test data explores a different region of the input than in the training data. Further, learning of complex models from inputs might be difficult because shifting of input by bias could make saturation of neurons (value $\le$ 0) difficult.

\subsubsection{Summary of discussions on ICNN}
\begin{enumerate}
    \item ICNN performs well when used to learn convex functions. In such cases, optimization over ICNN to minimize an output is advantageous as a convex optimization  problem is  solved.
    \item When ICNN is used for non-convex functions, approximation errors may be quite high. In particular, the global optimum identified by ICNN need not necessarily be close to true global optimum or true local optimum.
    \item Recurrent ICNN cannot model time delayed inputs at the output in the absence of bias. In the presence of bias, the performance could be attributed to the shifting of the inputs, which might result in poor generalization.
\end{enumerate}

\section{ CDiNN- Convex Difference Neural Network}

We now describe a neural network architecture based on the idea of modeling arbitrary nonlinear functions as a difference of polyhedral convex functions. We call this structure the CDiNN (Convex Difference Neural Network) since this network learns the output as difference of convex functions from its constituent ICNN units. In this section, we start with a brief description of difference of convex (DC) functions and their properties. We then focus on a method called the Convex Concave Procedure (CCP) \cite{CCP} used in the optimization of DC functions. The properties of this network structure and extensions to a recurrent form are described.

\begin{figure}
	\centering
		\includegraphics[scale=.5]{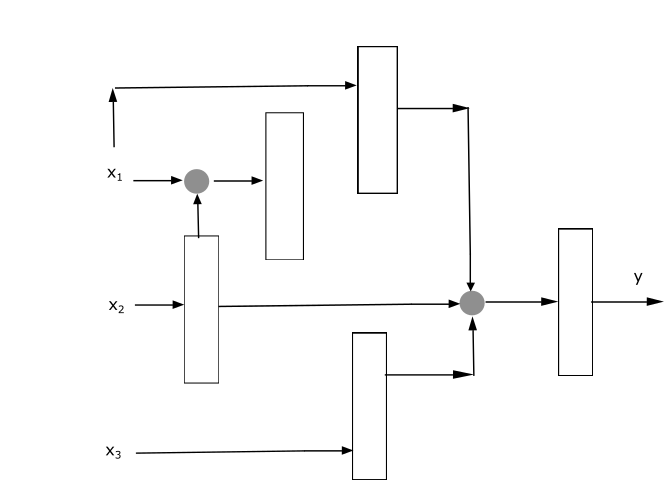}
	\caption{Psuedo multilayer network. Each box is one Unary single layer network and grey circles are summations}
	\label{FIG:psuedo}
\end{figure}

Before we proceed, we point out one attempt \cite{azutd94244:online}, where the use of DC optimization with neural networks was discussed. In this work, a given function is decomposed as unary functions and each unary function is represented as a single hidden layer network. A library of unary functions is described. With this representation and the decomposition scheme, the original function can be represented as a "pseudo multilayer neural network" (Figure \ref{FIG:psuedo}). This multilayer network can be rewritten manually as a DC function. A few example functions are used to demonstrate this idea  in this thesis. However, the focus of the work was not on learning from data but more on exploring the representational capability. Even from a representational viewpoint, the procedure is not automatic. Nonetheless, it is worthwhile to point out that researchers had identified the potential use of DC functions in neural networks.

In another work, \cite{sivaprasad2020curious} describe an ICNN architecture for binary classification, where the output is a two dimensional vector. Each element of this vector is a convex function of the inputs. Argmax of this output vector predicts the class of the input. This results in a decision boundary, which is a difference of convex function. Further, they explore an ensemble of ICNNs for multi-class classification. A gating layer allows the choice of ICNN to be used based on the region of interest. However, they do  not explicitly formulate a difference of convex neural network architecture. As a result, the use of difference of convex optimization techniques cannot be applied to find optimal inputs.  Further, the use of the difference of convex structure for regression and optimization are not discussed. Additionally, we believe that the use of difference of convex optimization approaches will be of critical importance in further development of training approaches for such neural network structures.

\subsection{Difference of Convex functions and properties}
Difference of Convex (DC) functions are those class of functions in which the output is represented as a difference of convex functions of inputs as in Equation \ref{eq3.2}. In this equation, $f_1$ and $f_2$ are convex functions of input x. Many nonlinear functions can be represented as Difference of Convex (DC) functions \cite{12funcdc}. For example, theorem \ref{th1} holds good for any twice differentiable function. A convex function that is represented as a maximum of finite number of affine functions is said to be a polyhedral convex function. A DC function where either or both of the convex functions is(are) polyhedral convex function(s) is said to be a difference of polyhedral convex functions.
\begin{equation}
\label{eq3.2}
f(x) = f_1(x) - f_2(x)
\end{equation}

\begin{theorem}
\label{th1}
Any twice differentiable function $f:x->y$ can be represented as a difference of two convex functions. \cite{CCP}
\end{theorem}

\subsection{Difference of Convex (DC) optimization}
\label{DCopt}
DC Optimization deals with minimizing an objective function with constraints, when both the objective and constraints are represented by difference of convex functions. There are different algorithms that solve DC optimization problems. Here, we discuss a specific algorithm called Convex-Concave Procedure (CCP) which can be considered as a version of Difference of Convex Algorithm (DCA) \cite{CCP}. DCA involves solving alternating sequences of primal and dual problem whereas CCP solves the problem directly as a primal formulation. More details on comparison between DCA and CCP and their extensions can be found in \cite{CCP}. 
Consider the DC optimization problem,
\begin{equation}
\begin{array}{ll} 
&{\text{ minimize}}\quad f_0(x)- g_0(x) \\ 
&{\text{ subject}}\,{\text{to}}\quad f_i(x) - g_i(x) \le 0, \quad i = 1,\ldots ,m, \end{array}
\end{equation}

CCP is an iterative procedure that involves linearization of the concave part of the objective function ($-g_0(x)$) and constraints ($-g_i(x)$) to get a convex problem at each iteration. Then this convex problem is solved to global optimum to get an initial point for the next iteration. The linearization used at each iteration is a first order Taylor's series expansion around the point $x_0$, $g_0(x) = g_0(x_0) + \nabla (g_0(x_0)) \cdot (x - x_0)$, where $x_0$ is the initial guess or the optimum value of x from the previous iteration. This algorithm converges to a local optimum. Further, it is noted that at each iteration, CCP is guaranteed to produce a better result than the previous iteration. This is because the convexified function $ \hat{f}(x) \ge f(x)$, since linearization of a concave function provides an upper bound to that function. This means that the value of the true objective function is always less than the value of the modified objective. This result helps establish convergence as explained in detail in \cite{CCP}.

In particular, it is easy to observe that if the DC function is a difference of polyhedral convex functions with $f_0(x)$ in the form of maximum of affine functions and when constraints are either absent or affine, then the solution to the convex problem at each iteration is a Linear Programming (LP) problem as given below. Further, we note that some DC optimizations methods have nice properties when the convex functions in DC representation are polyhedral. For example, DCA has finite time convergence guarantee for difference of polyhedral convex functions. \cite{dc11}, \cite{polyhedra}

\begin{flalign}
x_{opt} = argmin ( max (A^1x + b^1, A^2x + b^2 ... A^nx + b^n))
\end{flalign}
This can be written as linear problem as:
\begin{flalign}
x_{opt} = argmin ( t) 
\end{flalign}
with the constraints,
\begin{flalign*}
A^1x + b^1 \le t \\
A^2x + b^2 \le t\\
...\\
A^nx+b^n \le t
\end{flalign*}

\subsection{Design of CDiNN: Convex Difference Neural Network}
We propose a Neural network architecture called Convex Difference Neural Network (CDiNN) where the output is learnt as a difference of polyhedral convex functions of the input. We discuss two ways in which CDiNN can be designed in this section, which are both based on the ICNN structure. Before that, we discuss the necessity of pass-through layers in the ICNNs and investigate options to remove them.

\subsubsection{Pass-through layers}
\label{pathrough}
\begin{figure}
	\centering
		\includegraphics[scale=.6]{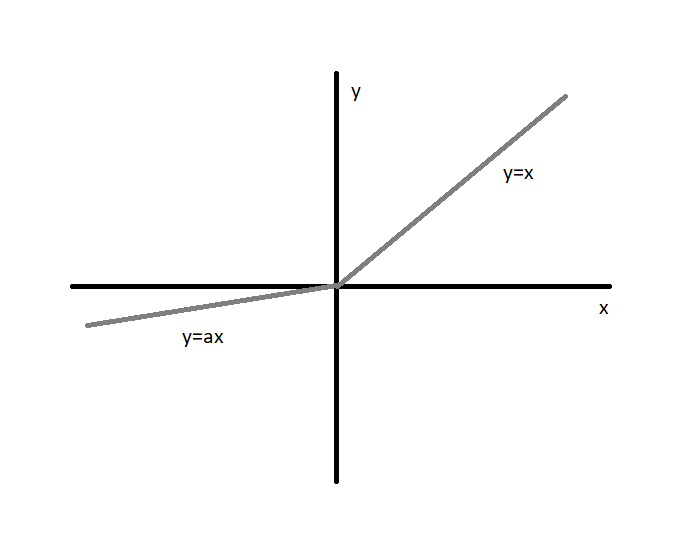}
	\caption{Parameter-Constrained ReLU (PC-ReLU) output = max $( ax, x)$. This function is convex non-decreasing as $a \in [0,1]$ }
	\label{FIG:pararelu}
\end{figure}

Consider ICNN with a linear activation function at the output layer. In the absence of pass-through layers, ICNN learns the function as a maximum of affine representations, that is, $max (A^1x + b^1, A^2x + b^2 ... A^nx + b^n, 0) + b^{n+1}$ where $b^{n+1}$ is the bias of output layer. This is because ReLU activation function clips the negative values at each hidden layer. In the presence of pass-through layers, the pass-through at the output layer generates an additional affine term that renders the output to $max ((A^1x + b^1, A^2x + b^2 ... A^nx + b^n, 0) + (A^{n+1}x + b^{n+1}))$.  Hence removal of pass-through layers restricts the class of functions that can be learnt by ICNN. In particular, when biases are ignored, even identity mapping cannot be learnt by ICNN without pass-through layers. One approach to address this problem is suggested in \cite{sivaprasad2020curious}, where it is proposed to use Leaky ReLU or ELU as activation function. ELU activation function is given by output = ($\alpha \cdot (e^x-1)$ if $x<0$, and $x$ if $x>=0$) and Leaky ReLU is given by output = ($\alpha x$ if $x<0$, and $x$ if $x>=0$), where $x$ is the input and $\alpha$ is a hyperparameter. In that work, results are generated using ELU activation. It could be seen that with the use of ELU activation function, convexity with respect to inputs is retained as long as $\alpha \le 1$.
 We propose the use of Parametric ReLU as shown in Figure \ref{FIG:pararelu} \cite{6deeprect}. The activation function is represented by output = ($ax$ if $x<0$, and $x$ if $x>=0$), where $x$ is the input. It has one learnable parameter '$a$'. Parametric-ReLU is similar to leaky ReLU, except that instead of choosing a small value for $a$, this parameter is learnt as a part of network training. If we constrain this learnable parameter to be non-negative, then Parametric-ReLU is a convex non-decreasing function. We can represent this function as $max(ax,x)$ with $a \in [0,1]$. We call this as Parameter-Constrained ReLU (PC-ReLU).  Thus, the use of PC-ReLU allows for negative values as inputs in the inner hidden layers without affecting the convexity.
 
 It can be seen that PC-ReLU acts as a simple pass-through if $a=1$. As a result, if the dimension of the input is m and if there are at least m neurons at each hidden layer, the network could learn the activation functions and weights such that the inputs are passed through to hidden layers. From Theorem \ref{th2} and remark \ref{themax2}, we see that ICNN with one neuron and a pass-through connection at each hidden layer and ReLU activation function can approximate any Lipschitz convex function. Hence, ICNN without pass-through layer and PC-ReLU activation function can approximate any Lipschitz convex function if there are at least $m+1$ neurons at each hidden layer and if the network has sufficient hidden layers. In essence,  PC-ReLU activation function permits the network to learn implicit pass-through layers whenever needed. It is not clear if ELU or Leaky-ReLU would have similar representational capability.

 \subsubsection{Architecture of Feedforward CDiNN}
\begin{figure}
	\centering
		\includegraphics[scale=.75]{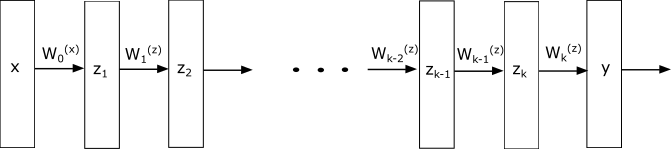}
	\caption{CDiNN-1 architecture. All the weights except $W_0^{(x)}$ ,$W_{k}^{(z)}$ are constrained to be non-negative}
	\label{FIG:2.1}
\end{figure}

\begin{figure}
	\centering
		\includegraphics[scale=.75]{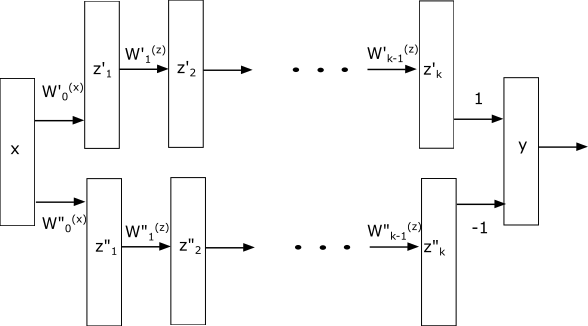}
	\caption{CDiNN-2 architecture. All the weights except $W_0^{'(x)}$, $W^{''(x)}_{0}$ are constrained to be non-negative}
	\label{FIG:2.2}
\end{figure}

Simple extensions of the ICNN architecture allows the design of neural networks that can learn functions as difference of convex functions. We observe that it would be sufficient to relax the non-negativity constraint at the last layer in an ICNN to produce a difference of convex mapping. If $z_k$ represents the outputs of neurons in the last layer in an ICNN, every mapping in $z_k$ layer is convex with respect to the inputs. Knowing that non-negative sum of convex functions is convex and a negated convex function is a concave function, we can realize a CDiNN structure by adding an output layer after $z_k$ layer in the ICNN structure. In this case, the connecting weights from $z_k$ layer to this output layer are unconstrained,

\begin{flalign}
\label{eq3_4}
y &= f(z_k) \\
\label{eq3_5}
y &= \sum (W^{p(z)}_k z_{kp}) + \sum(W^{n(z)}_k z_{kn})
\end{flalign}

where, $W^{p(z)}_k$ are the positive weights and $W^{n(z)}_k$ are the negative weights connecting the last layer $z_k$ to output layer. This formulation can be considered as sum/difference of outputs of multiple ICNNs. This is shown in Figure \ref{FIG:2.1}. We denote the CDiNN architecture which has a single neural network that follows equation \ref{eq3_5} as CDiNN-1. In the second extension, we propose two ICNNs without interconnections in hidden layers whose outputs are subtracted at the final layer as shown in Figure \ref{FIG:2.2}.  This CDiNN architecture with two independent ICNNs as CDiNN-2. We note that  CDiNN-1 configuration has lesser number of nodes in hidden layer in comparison to CDiNN-2 for the same number of network parameters and depth. This is due to the absence of interconnections in the hidden layer between two ICNNs in CDiNN-2. 

\subsubsection{Pass-through layers and choice of activation function}
\label{Relu}
We showed that through the use of PC-ReLU, pass-through layers might not be needed in ICNN structure that forms the basis for CDiNN.  We described in section \ref{pathrough} that ICNN with ReLU activation function clips the negative values from the affine functions when there are no pass-through layers and biases are disabled. However, when these ICNNs are used in CDiNN, the convex functions are subtracted which permits negative values of affine functions to be represented. This shows that after the removal of pass-through layers, CDiNN has better representational ability than ICNN even if ReLU is used as an activation function. Not withstanding this, in this paper, we use PC-ReLU as the activation function for CDiNN.

\subsubsection{Design of Recurrent CDiNN}
\begin{figure}
	\centering
		\includegraphics[scale=0.75]{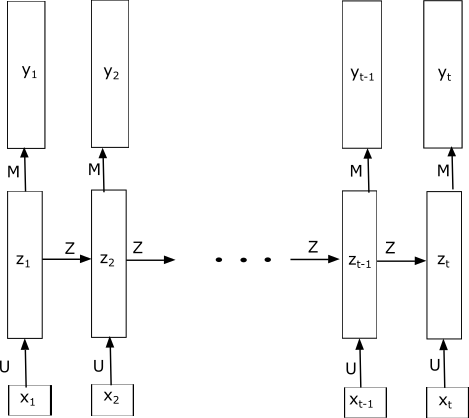}
	\caption{Recurrent CDiNN}
	\label{FIG:drnn}
\end{figure}

We extend the recurrent ICNN architecture to design recurrent CDiNN as follows. With the removal of pass-through layers and choice of PC-ReLU as activation function, recurrent convex functions in the recurrent CDiNN is modeled by equation \ref{rcdinn}, where weights Z are non-negative. As M is unconstrained, $y_t$ is in the DC form. The recurrent CDiNN is shown in Figure \ref{FIG:drnn}.

\begin{eqnarray}
z_t & = & \textrm{PC-ReLU}( U x_t + Z z_{t-1}) \nonumber \\
y_t & = & ( M z_t) \label{rcdinn}
\end{eqnarray}
 
\subsection{Properties of CDiNN}
\begin{enumerate}
    \item \textbf{Representation capability}.
    Any function that is capable of being decomposed into difference of convex functions can be learnt by CDiNN. 
  \noindent
We find from Theorem \ref{th2} that any convex function can be learnt by ICNN. Since CDiNN learns the output as difference of outputs of ICNNs, we see that the structure of CDiNN is capable of learning of any function that can be written as difference of two convex functions.
We find from Theorem \ref{th1} that any twice differentiable function can be represented as difference of convex functions. Thus, it is can be seen that any twice differentiable function can be learnt by CDiNN.

\begin{remark}
It is observed that any neural network with one hidden layer of ReLU activation function and linear output layer is a CDiNN Network. This is because at the output of hidden layer, each function is convex and the output is a weighted sum and can be written as a difference of convex functions as in Equation \ref{eq3_5}.
\end{remark}

\item \textbf{CDiNN learns the function as difference of polyhedral convex functions}
We note that ICNN learns a function in the form of maximum of finite number of affine functions. Since CDiNN learns the function mapping as difference of outputs of multiple ICNNs, the function represented by CDiNN is in the form of difference of polyhedral convex functions.

\end{enumerate}

\subsubsection{Training of CDiNN}
In this paper, a modification to the training procedure used for ICNN is pursued while training the CDiNN architecture. This removes the step that heuristically forces the weights to be non-negative in ICNN training. To constrain the weights that connects layers to be non-negative, we modify the forward function by use of squared parameters. In the context of this paper, in all the examples used in the next section on discussions and applications, $(W_i^{(z)})^2$ are used as weights in the hidden layers of CDiNN. Adam optimizer \cite{kingma2017adam} with Xavier-normal initialization \cite{xavier} of weights and zero initialization of biases are used in training all the neural networks described hereon.

\section{Discussions and Applications}
In this section, the properties of CDiNN presented previously are explored. First, we empirically show the representational capability of CDiNN. We then discuss the results of optimization with CDiNN and compare it with ICNN and Standard ReLU neural networks. Finally, we present one simple engineering example to highlight the advantages of the CDiNN structure. In this first paper on CDiNN, the intent of these examples is to describe the fundamental aspects of this architecture in detail. Benchmarking studies on large engineering examples and theoretical and application studies on optimizing the network structure for performance will be pursued in the future.

\subsection{Effect of removing Pass-through layers}
We compare the performance of ICNN and CDiNN in the absence of pass-through layers. We discussed in Section \ref{pathrough} that ICNN with ReLU activation function would not even be able to learn identity mapping $y = f(y)$ in absence of pass-through layers when biases are ignored. But when such ICNNs are used in CDiNN, convex functions could be learnt as $f_1 = ReLU(x)$ and $f_2 = ReLU(-x)$ and the output can be learnt such that $output = f_1 - f_2 = x$ . This shows that in the absence of pass-through layers, CDiNN is better at learning even convex/affine functions in comparison to ICNN. We further show empirically in Table \ref{tbl1} (through simple nonlinear functions) that the removal of pass-through layers in the individual ICNNs in the CDiNN structure does not significantly impact the representational capability.

\begin{table}[width=.9\linewidth,cols=4,pos=h]
\caption{Fit MSE for 1D and 2D synthetic function on CDiNN network with and without Pass-through layers}\label{tbl1}
\begin{tabular*}{\tblwidth}{@{} LLLL@{} }
\toprule
 Function & Fit MSE with Pass-through layer &  Fit MSE without Pass-through layer \\ 
\midrule
 1D Sine & 0.0007 & 0.0003  \\ 
 1D Quadratic &  0.0024  & 0.0016 \\  
 1D Cubic & 0.0032 & 0.0021   \\  
 2D Circles & 0.069 & 0.069  \\ 
 2D Moons &  0.0385 & 0.0375  \\   
\bottomrule
\end{tabular*}
\end{table}

\subsection{Non-linear function approximation}
Synthetic 1D and 2D functions are used to show that CDiNN is capable of learning nonlinear functions.
\subsubsection{1D Regression}
\begin{figure}
	\centering
		\includegraphics[scale=.4]{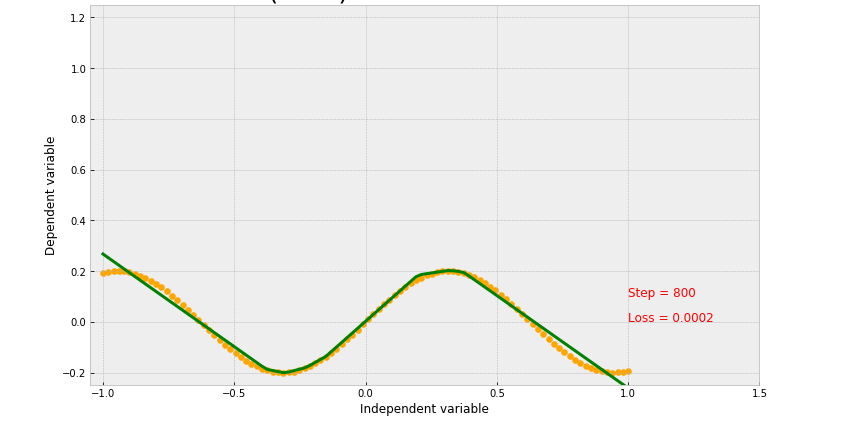}
	\caption{Learning of sine by CDiNN. The green curve is the learnt function, the yellow points represent the training data. Mean square Loss of training data is 0.0002 and number of training epochs  is 800.}
	\label{FIG:4}
\end{figure}

\begin{figure}
	\centering
\includegraphics[scale=.3]{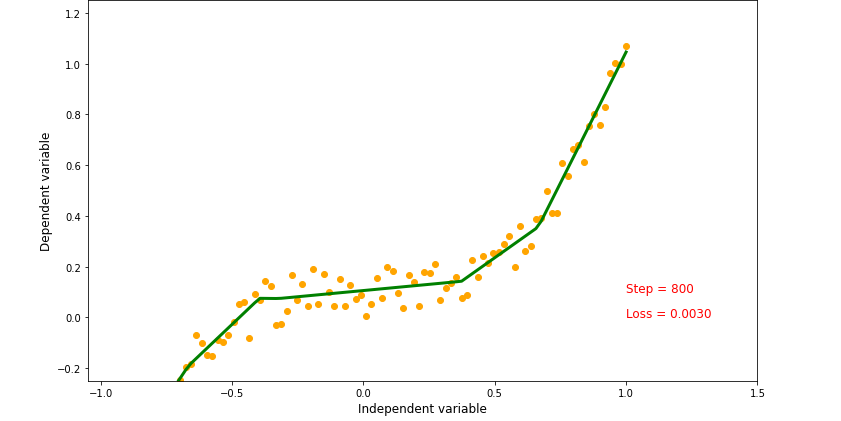}\includegraphics[scale=.3]{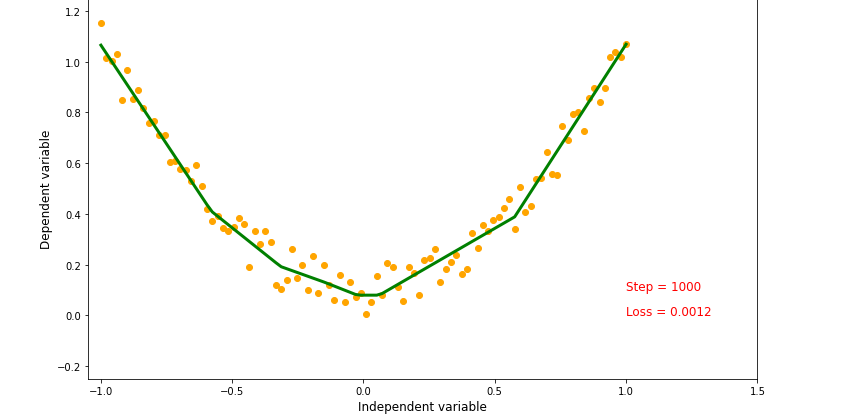}
	\caption{Learning of quadratic and cubic functions by CDiNN . The green curve is the learnt function, the yellow points represent the training data. Steps and loss indicate number of training epochs (800) and Mean square loss of training data respectively}
	\label{FIG:5}
\end{figure}

Consider the non-convex functions: $y = sin(5x)/5$, $y=x^{2}$ and $y=x^{3}$ where, $x \in [-1, 1]$. The fit of the function for CDiNN Network is shown in Figures \ref{FIG:4} and \ref{FIG:5}. We note that CDiNN is able to model all the three functions with reasonable accuracy as can be seen from Table \ref{tbl1}.

\subsubsection{2D classification}

\begin{figure}
	\centering
		\includegraphics[scale=.25]{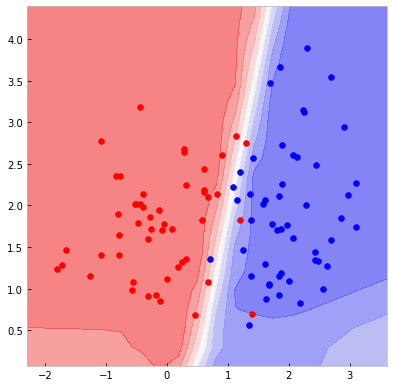}\includegraphics[scale=.25]{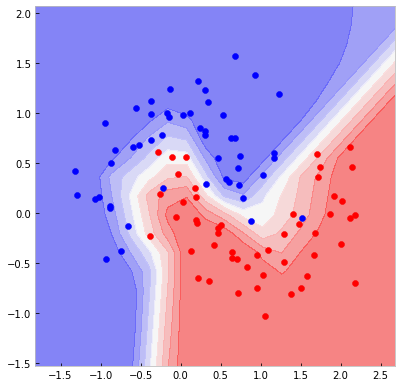}\includegraphics[scale=.25]{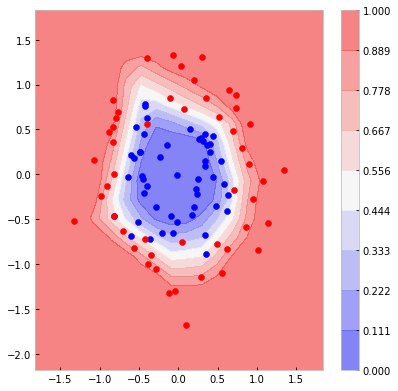}
	\caption{Classification decision boundaries of CDiNN  on 2D synthetic data}
	\label{FIG:6}
\end{figure}

In Figure \ref{FIG:6}, classification performance of CDiNN on a 2D dataset is illustrated using a two-dimensional binary classification problem from the sci-kit-learn toolkit \cite{4scikit-learn}. The points in the picture represent training data with output as either 0 or 1 and the regions are coloured based on the value of predictions $\in [0,1]$. It could be seen that the CDiNN is able to learn complex decision boundaries.

\subsection{Time delay system}
\label{tds}
\begin{figure}
	\centering
		\includegraphics[scale=.25]{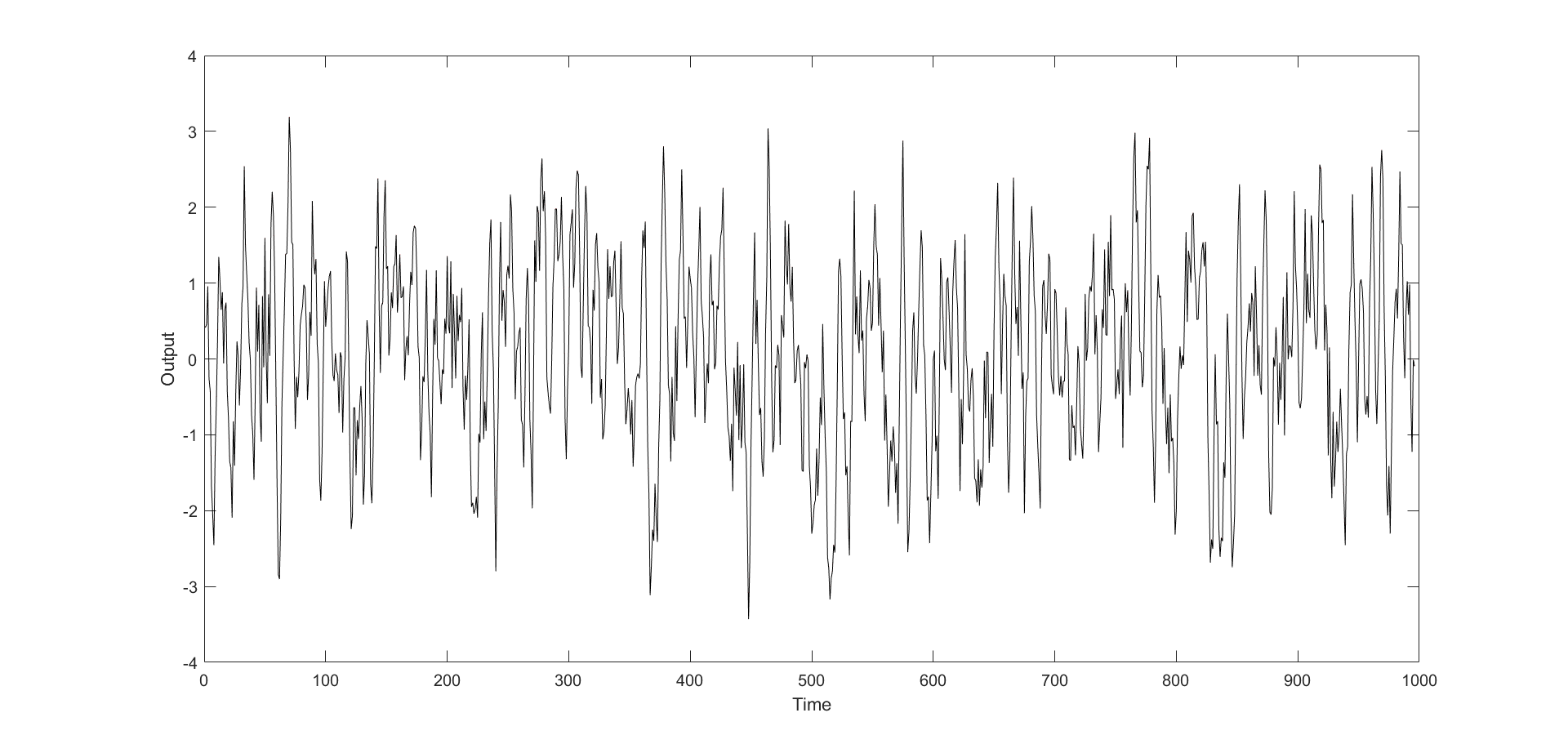}
	\caption{Output predicted (red) vs truth (black) by CDiNN RNN. The red line and black curves almost coincide with each other}
	\label{FIG:7}
\end{figure}

In section \ref{tds1}, we saw that the recurrent version of ICNN cannot  model time delay systems effectively.  We now proceed to show that CDiNN network can model time-delay systems. Assume, $y_t = u_{t-m}, m \ge 0$. Let PC-ReLU parameter be set to 0 so that activation function is ReLU. We see that the network could learn $y_{t_1} = ReLU(u_{t-m})$ and $y_{t_z} = ReLU(-u_{t-m})$. Hence $y_t = y_{t_1} - y_{t_2} = u_{t-m}$ can be learnt. We also discuss the experimental results on the response of CDiNN with 10 recurrent neurons. This has similar number of learnable parameters to ICNN discussed earlier. In this case as well, we disable the learning of bias in recurrent and feed-forward layers to avoid effects due to shifting of input. The input that is used as excitation signal is $u_t = \textrm{randomUniform()} \in [-1,1]$, which is random signal uniformly sampled from [-1,1]. The output is $y_{t} = u_{t-4} + u_{t-3} + u_{t-2} + u_{t-1}$ and the number of samples given per sequence as inputs to the network is 5. We see that the CDiNN is able to capture the pattern of noise in the input at output $y_t$ from Figure \ref{FIG:7}.

\subsection{Discussions on Optimization}
We now demonstrate the utility of CDiNN in solving standard optimization functions. We first learn the optimization function in DC form using CDiNN and use CCP to solve it. We compare the results of optimization to ICNN and Standard-ReLU neural networks.
\subsubsection{Test setup}
We use three optimization test functions from \cite{simulationlib}. We add a constant value of 5 to each of the functions. We train the network by using scaled inputs and outputs (so that inputs and outputs $\in$ [-1,1]).
\begin{itemize}
    \item \textbf{$ f = $ Three humped camel function $ + 5$}: This is a Non-convex 2D input function. Global minimum is 5. The maximum value of function in training set is 2.1E+03, which is scaled to 1 for training the network.
    \item \textbf{$ f = $  Sumpower $ + 5$}: This is a convex 5D input function. True minimum is 5. The maximum value of function in training set is 10, which is scaled to 1 for training the network.
    \item \textbf{$ f = $  Matya$ + 5$}: This is a convex 2D input function. True minimum is 5. The maximum value of function in training set is 105, which is scaled to 1 for training the network.
\end{itemize} 
  The experiment involves training the following networks.
\begin{itemize}
    \item Standard Parametric-ReLU network of 30 hidden layer neurons
    \item CDiNN-1 network with 30 hidden layer neurons
    \item CDiNN-2 network with 15 hidden layer neurons in each of its ICNNs
\end{itemize} 
All the networks have only one learnable hidden layer and almost equal number of learnable network parameters.

After training, we optimize the input for minimizing the output. The initial value for optimization was [-1,-1] for SumPower function and [1,1] for Matya and Camel function. 
The optimization used for standard ReLU network is to use Filtered-$\beta$ (sub-)gradient descent \cite{gd} \cite{d2}. The algorithm is stopped based on a certain convergence criterion or after a maximum number of iterations. At every iteration k,
\begin{eqnarray}
s_k & = & (1-\beta)\cdot g_k + \beta \cdot s_{k-1} \nonumber \\
x_{k+1} & = & x_k - \alpha_k \cdot s_k \nonumber
\end{eqnarray}
where $\alpha_k$ is step size at iteration k, $\beta$ is set to 0.25 and $g_k$ is the gradient of output with respect to input at iteration k.
The step size used is either Constant step size ($\alpha = constant$) or  "Square summable but not summable" step size ($\alpha = \alpha/k$). The experiments are run with different starting step sizes $[0.01,0.1,5,10]$. The optimization for CDiNN network is performed using CCP procedure as explained in Section \ref{DCopt}. Standard linear programming optimizers from CVXOPT Modelling \cite{CVX99} library in python are used to find the optimum value of the convexified function at each iteration. The number of maximum iterations is set to 200 or 500. The stopping criteria for convergence is that the difference of function values between successive iterations is either $1e^{-3}$ or $1e^{-5}$. Since training of the non-convex neural network can lead to different parameters due to local optima, we train the network three times and report the results as follows. Choose the network that reported best optimization results in CCP for CDiNN-1 and CDiNN-2 and choose the network that reported best optimization results in Gradient descent for Standard ReLU network. This is to done to ensure that results of optimization are not affected due to local optima problems in training the neural networks. \footnote[1]{The time taken for optimization needs to be understood with caution since the algorithm is not run on dedicated processors and these might not be the best implementations of the corresponding optimization algorithms.}

\begin{table}[width=.9\linewidth,cols=7,pos=h]
\caption{Fit and optimization performance for Optimization test functions with stopping precision = 0.001}\label{tbl2}
\begin{tabular*}{\tblwidth}{@{} LLLLLLL@{} }
\toprule
 Func. &  Type &  Algorithm & Fit MSE &  Optimization parameters & $y_{opt}$ &  Exec. time (s) \\
\midrule
Camel &  CDiNN-1& CCP & 3.26-05&  & 5.07  & 0.04 \\ 
Camel &  CDiNN-2 & CCP  & 1.9E-05 & &  6.97 & 0.041 \\

Camel & Std. ReLU & GD  & 3.7E-05 &$\alpha_0 = 0.1$, $\alpha_{n+1}=\alpha_n$ &  8.31 & 3.69 \\
Camel & Std. ReLU & GD  & 3.7E-05 & $\alpha_0 = 5$, $\alpha_{n+1}=\alpha_n$&  2001.46 & 1.37 \\

Sum power &  CDiNN-1 & CCP  &  1.29e-05 &  &  5.02 & 0.07 \\ 
Sum power &  CDiNN-2 &  CCP & 8.2E-04 &  &  5 & 0.06 \\
Sum power &  Std. ReLU & GD & 8.7E-05 &  $\alpha_0 = 5$, $\alpha_{n+1}=\alpha_n$ &  5 & 6 \\
Sum power &  Std. ReLU & GD & 8.7E-05 &  $\alpha_0 = 0.01$, $\alpha_{n+1}=\alpha_n/n$ &  9.86 & 0.06 \\

Matya &  CDiNN-1 &  CCP & 7.8e-06 &  &  5.15 & 0.08 \\ 
Matya &  CDiNN-2 &  CCP & 1.8e-05 & &  5.23 & 0.08 \\

Matya &  Std. ReLU & GD & 4.1E-06 & $\alpha_0 = 10$, $\alpha_{n+1}=\alpha_n/n$ &  5.08 & 1.78 \\

Matya &  Std. ReLU & GD & 4.1E-06 & $\alpha_0 = 10$, $\alpha_{n+1}=\alpha_n$ &  104.80 & 4.62 \\

\bottomrule
\end{tabular*}

\end{table}

\begin{table}[width=.9\linewidth,cols=7,pos=h]
\caption{Fit and optimization performance for Optimization test functions with stopping precision = $0.00001$} \label{tbl2.1}
\begin{tabular*}{\tblwidth}{@{} LLLLLLL@{} }
\toprule
 Func. &  Type &  Algorithm & Fit MSE &  Optimization parameters & $y_{opt}$ &  Exec. time (s) \\
\midrule
Camel &  CDiNN-1 & CCP & 3.26-05&  & 5.07  &  0.04 \\
Camel &  CDiNN-2 & CCP  & 1.9E-05 &  &  6.97 & 0.04 \\
Camel &  Std. ReLU & GD & 3.7E-05 &  $\alpha_0 = 0.1$, $\alpha_{n+1}=\alpha_n$ &  8.31 & 3.69 \\
Camel &  Std. ReLU & GD & 3.7E-05 & $\alpha_0 = 10$, $\alpha_{n+1}=\alpha_n/n$ &  8.98 & 3.99 \\
Camel&  Std. ReLU & GD & 3.7E-05 & $\alpha_0 = 5$, $\alpha_{n+1}=\alpha_n$ &  2001.46 & 1.37 \\

\bottomrule
\end{tabular*}
\end{table}

\begin{figure}
	\centering
		\includegraphics[scale=.32]{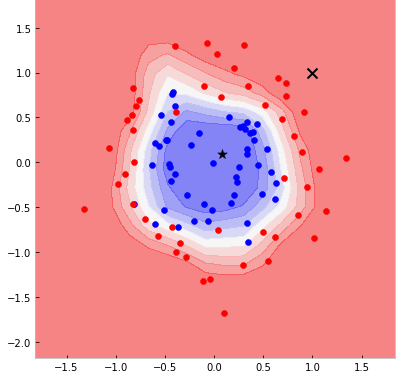}\includegraphics[scale=.32]{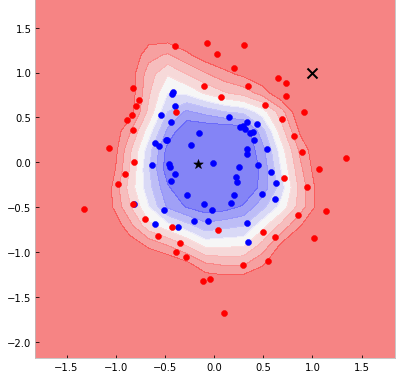}\includegraphics[scale=.32]{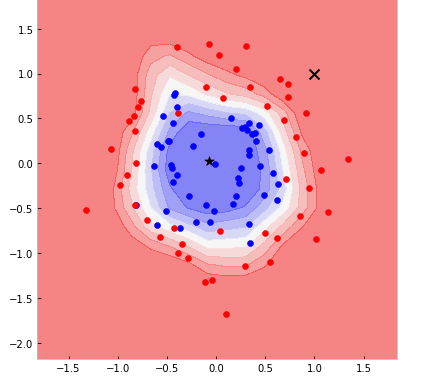}
	\caption{Optimization using CCP on the classification dataset. The initial points (cross) and the final solution (star) are marked for a) Standard NN b) ICNN c) CDiNN }
	\label{FIG:8}
\end{figure}

\subsubsection{Test result Discussion}
We discuss the performance of CDiNN and Standard ReLU neural networks from Table \ref{tbl2} and \ref{tbl2.1}.
\begin{enumerate}
    \item \textbf{Learning the function:}
    We observe that fit performances of CDiNN-1 and CDiNN-2 network are on par with Standard ReLU neural networks.  
    \item \textbf{Execution time: }Execution times of CDiNN-1 and CDiNN-2 are also comparable to each other with the available results. No conclusions are drawn based on execution times between gradient descent and CCP since the codes are not optimized versions of these algorithms. But we observe that the execution time of CCP optimization algorithm on CDiNN network shows little variation for a given neural network, which is ideal for real-time applications. In real time applications, it is desirable to have a fairly consistent execution time rather than unpredictable execution times.
    \item \textbf{Variability in results: }We see that CCP has no hyper-parameter tuning (except for choice of convergence precision) and the results from Standard ReLU network show a higher variance with respect to optimization hyper-parameters.
    \item \textbf{Optimization results: }In general, we see that CDiNN-CCP pair produces results that are very close to the optimally tuned gradient descent algorithm. However, the tuning parameters in the CDiNN-CCP solution have very little effect on the final solution from the viewpoints of accuracy and execution  time.
\end{enumerate}

\subsubsection{Finding partial inputs}
\begin{figure}
	\centering
		\includegraphics[scale=.32]{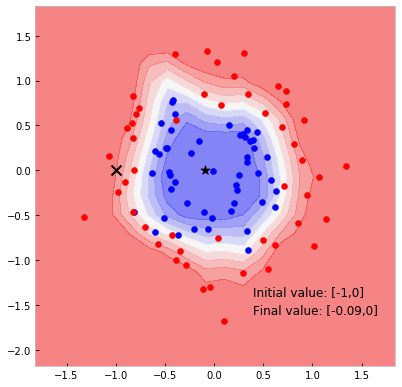}\includegraphics[scale=.32]{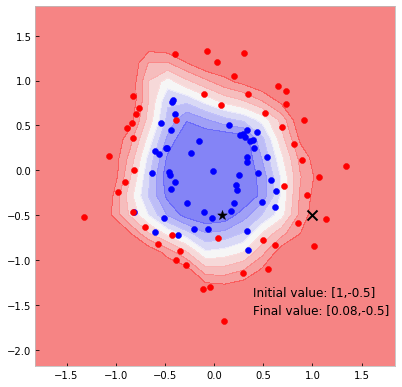}
	\caption{Partial input optimization using CCP on the classification dataset along horizontal axis. The initial points (cross) and the final solution (star) are marked}
    \label{FIG:9}
\end{figure}

Now we apply CCP in finding partial inputs. For illustration, we fix one of the coordinates (y coordinate) in the circles 2D classification task and run the optimization to find the other coordinate value that minimizes the output. The results are given in Figure \ref{FIG:9}.

\subsection{Estimating time and location of Pollutant spill}
In this section, we compare the performance of ICNN, CDiNN and Standard ReLU Neural network in estimating the time and location of pollutant spill from simulated data. We use the non-smooth function in \cite{simulationlib} for simulation which captures scaled pollutant concentration as a function of distance and time.

\subsubsection{Problem description}
The problem under study involves two-spills, first spill at location = 0 and time = 0 and the second spill at location = 0.8 and time = 10 and the system is simulated for $location \in [0.01, 1]$ and $time \in [0.01, 15]$.

Given the pollutant concentration measurements at some locations at different times, we would like to compare the fit of different neural networks and optimization to identify the location and time of the spill. The data for training the neural network is sampled at an interval of 0.02 along each dimension (location and time) and for test data both the inputs are sampled at an interval of 0.05. A weight of (-1) is added at the output of ICNN to enable it learn the output as concave function. The inputs are normalized to [-1,1] before training. Optimization is performed using (sub-) gradient ascent for standard ReLU, standard convex optimizers in python \cite{CVX99} are used for ICNN neural networks, and CCP algorithm is used on negated CDiNN output to determine the location and time of the spill. We also add the constraint that the feasible solution should be such that distance > 0.35 and time > 5.25. We use the starting point as d=1.2 and t = 15.2 and the optimum local solution is at d=0.8 and t=10.02. The optimal solution for time is considered as 10.02 seconds because the spill happens immediately after 10 seconds and the training data resolution is 0.02 seconds. We run the experiment two times and report the best optimization result in each network. This helps in reducing the effects of local minima in training the neural networks to affect the optimization results on the inputs.

\subsubsection{Test result discussion}

\begin{figure}
	\centering
		\includegraphics[scale=.3]{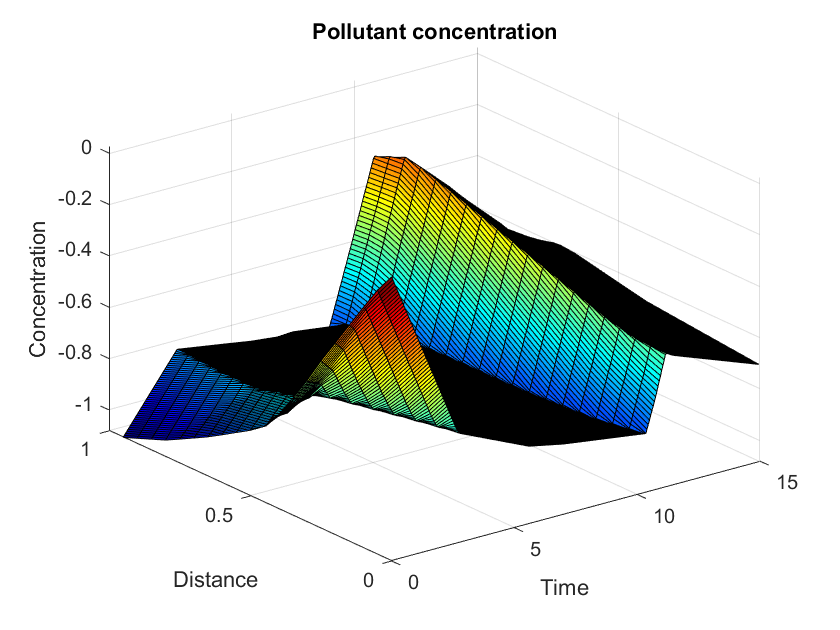}\includegraphics[scale=.3]{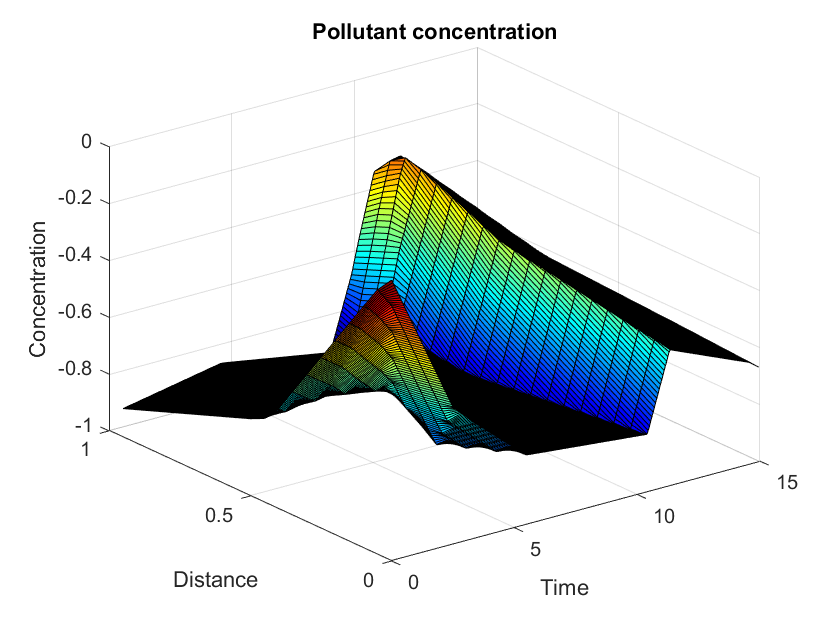}
	\caption{(a) The scaled model response captured by Standard Parametric-ReLU neural network (b) The scaled model response captured by CDiNN-2}
    \label{FIG:11}
\end{figure}

\begin{enumerate}
    \item \textbf{Learning the function:}
    The fit performance is shown in Table \ref{tbl3}. The performance of ICNN, CDiNN with the comparable number of learnable parameters is shown in Figure \ref{FIG:11}. 
    Further, the following observations show the ability of CDiNN to learn function mapping similar to standard ReLU networks. The true function is a non-smooth (not twice-differentiable) and discontinuous function at the location of second spill.  It is known that neural networks with continuous activation functions like ReLU wouldn't be able to model discontinuities. This is because ReLU is a continuous function and composition or summation of continuous functions is continuous. We see from Figure \ref{FIG:11} that the value of concentration at the point of discontinuity (concentration = 3.24) [d=0.8,t=10.02] is not learnt well in all the networks. But we observe that CDiNN is able to capture the function in difference of convex form with performances similar to that of Standard ReLU network.
    \item \textbf{Optimization: }
   The local maxima in time is expected immediately after 10 seconds. We consider time = 10.02 as the local solution since the training dataset is sampled at the resolution of 0.02 seconds.
   From Table \ref{tbl4}, we observe that ICNN performs poorly due to poor fit, the results from Standard ReLU NN-gradient ascent depends heavily on the choice of hyper-parameters but the optimization with CDiNN-CCP consistently provides the best results.
\end{enumerate}

\begin{figure}
	\centering
		\includegraphics[scale=.3]{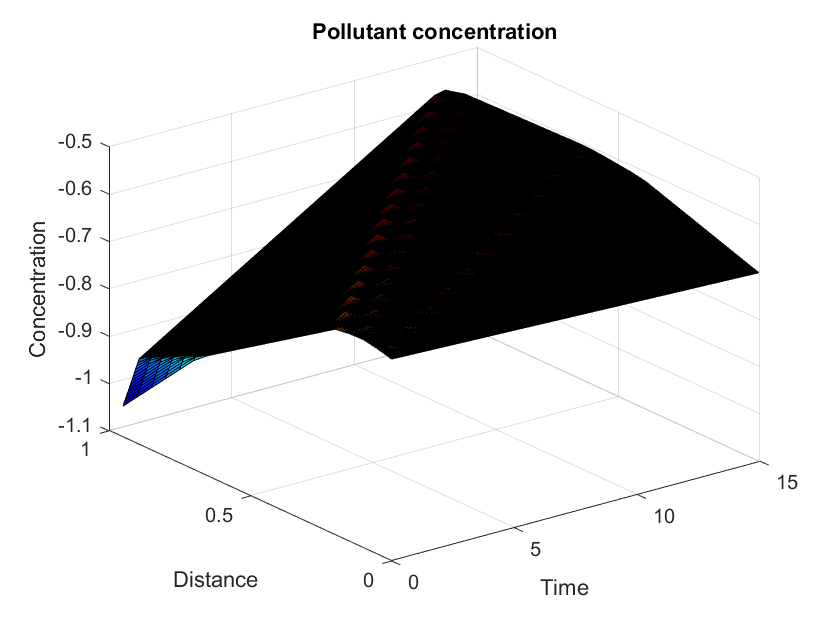}\includegraphics[scale=.3]{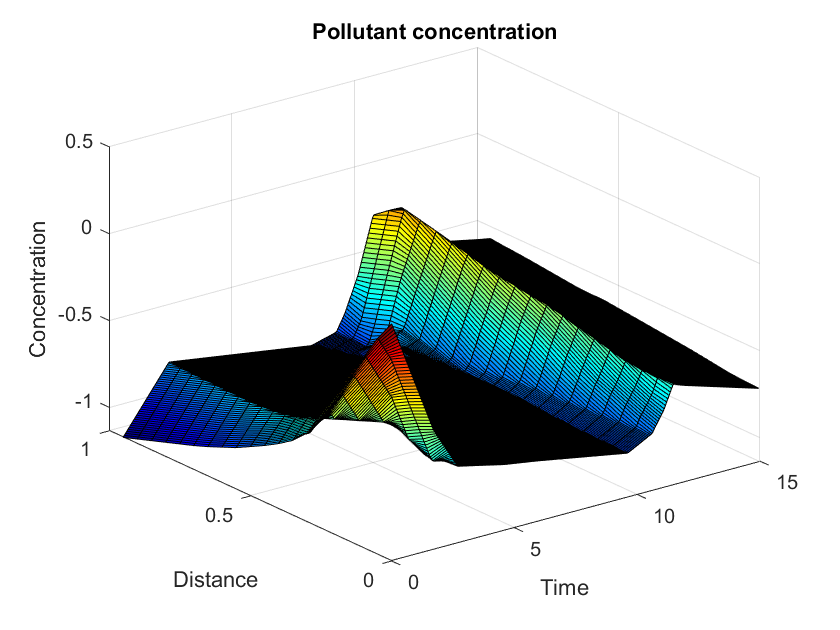}
	\caption{(a) The scaled model response captured by ICNN  (b) The scaled model response captured by CDiNN-1 }
    \label{FIG:10}
\end{figure}

\begin{table}[width=.9\linewidth,cols=2,pos=h]
\caption{Fit of different neural networks for Pollutant spill function mapping}\label{tbl3}
\begin{tabular*}{\tblwidth}{@{} LL@{} }
\toprule
 Network &  MSE of pollutant spill truth vs prediction \\
\midrule
 Standard Parametric-ReLU NN & 0.02 \\ 
 ICNN & 0.05 \\  
 CDiNN-1 &  0.02 \\  
 CDiNN-2 &  0.02 \\  
\bottomrule
\end{tabular*}
\end{table}

\begin{table}[width=.9\linewidth,cols=5,pos=h]
\caption{Optimization performance for identification of location and time of pollutant spill}\label{tbl4}
\begin{tabular*}{\tblwidth}{@{} LLLLLL@{} }
\toprule
 Title & Initial value $x_o$ & Step size & Final value $x_{opt}$ & $y_{opt}$ & $t_{exec}$ (seconds) \\ 
\midrule
 Expected local maximum &  d=0.8 t=10.02 & - & -  & 3.24 & -\\
 Standard Parametric-ReLU NN & d=1.2, t=15.2 & 0.01 & d=0.87 t=10.26 & 0.24 & 0.2\\
 Standard Parametric-ReLU NN & d=1.2, t=15.2 & 10 & d=0.57 t=5.25 & $-0.77$ & 0.21 \\
 Standard Parametric-ReLU NN & d=1.2, t=15.2 & 0.1 & d=0.92 t=11.86 & $-0.43$ & 0.2 \\
 ICNN & d=1.2, t=15.2 & - & d=0.91 t=12.95 & $-0.52$ & 0.02 \\
 CDiNN-1 & d=1.2, t=15.2 & - & d=0.84 t=10.23 & 0.35 & 0.04 \\ 
 CDiNN-2 & d=1.2, t=15.2 & - & d=0.87 t=10.31 & 0.16 & 0.03 \\ 

\bottomrule
\end{tabular*}

\end{table}

\section{Conclusion and future work}

In this paper, we discussed the disadvantages of ICNN and conclude that while ICNN provides a framework for capturing convex functions, use of ICNN for non-convex function approximation and subsequent use in control or any other application is not desirable since global optimum as learnt by ICNN might not be close to true global optimum. We also discussed that optimization results using (sub-)gradient descent methods on standard ReLU based feed-forward networks heavily depend on the choice of a number of hyper-parameters, with very little convergence guarantees. As a solution to these issues, we introduced a new neural network architecture called CDiNN. We showed that the use of Convex-Concave procedure with the new architecture  helps capture the non-linearity better and produces good quality local solutions with guarantees on convergence. We believe that the proposed network can be a good candidate to solve optimal control problems and other problems that involve finding the optimum input(s) to the neural network for a desired output. We anticipate that the application of global optimization techniques on CDiNN could lead to significantly improved results in several applications.

\bibliographystyle{cas-model2-names}

\bibliography{dc-refs}

\bio{Images/parameswaran_s_photo.jpg}
Parameswaran Sankaranarayanan is a doctoral research scholar at Indian Institute of Technology Madras. He does his research with Robert Bosch Centre for Data Science and Artificial Intelligence and Systems Engineering of Natural and Artificial group at Indian Institute of Technology Madras. He completed his bachelors in Electronics and Instrumentation Engineering in Madras Institute of Technology, Anna University Chennai. He has 7 years of industrial experience in automotive electronics and has done a number of projects involving embedded control systems.
\endbio

\bio{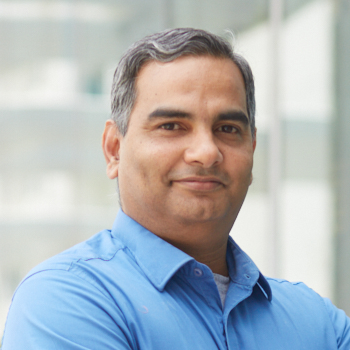}
Prof. Raghunathan Rengaswamy is a Professor at Indian Institute of Technology Madras. Prior to this, he was a Professor and co-director of the Process Control and Optimization Consortium (PCOC) at Texas Tech University, Lubbock, TX USA, Associate and full Professor at Clarkson University, Potsdam, NY and Assistant Professor at IIT Bombay, Mumbai, India.  Rengaswamy was the recipient of the Young Engineer Award for the year 2000 awarded by the Indian National Academy of Engineering (INAE). A paper that he co-authored was chosen by the International Federation of Automatic Control (IFAC) for the Best Paper Prize, for the years 2002-2005, in Engineering Applications of Artificial Intelligence Journal in the category – Application-oriented paper on Symbolic AI Approaches.  He was elected a Fellow of INAE in 2017.
\endbio



\end{document}